\documentclass{article}
\usepackage{spconf,amsmath,graphicx}
\usepackage{subfig}
\usepackage{booktabs}
\usepackage{multirow}
\usepackage{url}
\usepackage{enumitem}
\setlist{nosep, leftmargin=14pt}

\usepackage{mwe} % to get dummy images

\title{HemBLIP: A Vision--Language Model for Interpretable Leukemia Cell Morphology Analysis}
\name{Julie van Logtestijn$^{1}$ \qquad Petru Manescu$^{1}$}
\address{
    $^{1}$Department of Computer Science, University College London, United Kingdom\\
}
\begin{document}
%\ninept
%
\maketitle
\begin{abstract}

Microscopic evaluation of white blood cell morphology is central to leukemia diagnosis, yet current deep learning models often act as black boxes, limiting clinical trust and adoption. We introduce HemBLIP, a vision–language model designed to generate interpretable, morphology-aware descriptions of peripheral blood cells. Using a newly constructed dataset of $\sim$14k healthy and leukemic cells paired with expert-derived attribute captions, we adapt a general-purpose VLM via both full fine-tuning and LoRA-based parameter-efficient training, and benchmark against the biomedical foundation model MedGEMMA. HemBLIP achieves higher caption quality and morphological accuracy, while LoRA adaptation provides further gains with significantly reduced computational cost. These results highlight the promise of vision–language models for transparent and scalable hematological diagnostics.

\end{abstract}
\section{Introduction}
\label{sec:intro}

Leukemia remains a major global health concern, responsible for around 2.5\% of all new cancer diagnoses and 3.1\% of cancer deaths worldwide \cite{Sung2021}. Despite advances in therapy, the disease continues to impose substantial burden even in high-income countries, with an incidence of roughly 14 per 100,000 and a mortality rate near 6 per 100,000 per year \cite{SEERLeukemiaIncidence2025}. In low- and middle-income regions, survival rates can drop below 30\%, largely due to delayed or missed diagnoses linked to workforce shortages and limited laboratory infrastructure \cite{Bhakta2019,Sayed2015,DeAngelis2012}.

%Addressing this inequity requires not only therapeutic progress but also scalable, interpretable diagnostic tools that can operate where specialist expertise is scarce.

%Leukemia originates in bone marrow and disrupts normal hematopoiesis, producing abnormal white blood cells that proliferate uncontrollably.

Leukemia diagnosis relies on microscopical examination of peripheral blood smears (PBS) and bone marrow aspirates (BMA) by hematologists \cite{Sekar2023}. Morphological cues such as nuclear shape, chromatin texture, nucleoli, and cytoplasmic features are used to classify major subtypes—acute lymphoblastic (ALL), acute myelogenous (AML), chronic lymphocytic (CLL), and chronic myelogenous (CML) \cite{davis2014leukemiaoverviewprimarycare}. However, this process is labor-intensive, prone to human errors, and requires the availability of trained experts \cite{Amin2015}. 

%However, this process is labor-intensive and subjective: inter-observer variability can reach 10–20\% in blast counts, and manual microscopy error rates up to 40\% have been reported \cite{Dasariraju2020,Amin2015}. 
%These limitations highlight the need for computational methods that support early, accurate, and interpretable leukemia detection.

%Artificial intelligence (AI) has achieved promising results in hematological image analysis. Classical machine learning and convolutional neural networks (CNNs) can distinguish leukemic from normal cells with accuracies above 95\% on benchmark datasets \cite{Matek2019,Shafique2018,Dasariraju2020}.

Recent studies have demonstrated the potential of deep learning vision models to detect leukemic cells in PBS and BMA \cite{earlyALLdetection2024, yan2025diagnosis}. However, these task-specific classifiers often require frequent retraining when imaging conditions, staining protocols, or cell morphology distributions vary, which is difficult to sustain in real-world and low-resource settings \cite{Shorten2019,Lu2022}. More critically, their black-box outputs typically provide only class predictions without articulating the morphological reasoning behind them. This lack of interpretability limits clinical trust, validation, and ultimately adoption in routine hematology workflows \cite{Adadi2018,Holzinger2019}.

%old references for recent deep learning studies: Matek2019,Shafique2018,Dasariraju2020

%More importantly, their “black-box” nature and dependence on curated data have limited clinical adoption. Such models typically provide class labels without morphological reasoning, reducing trust and interpretability \cite{Adadi2018,Holzinger2019}. 

%Their performance equally degrades across laboratories or staining protocols, posing challenges in real-world and low-resource contexts \cite{Shorten2019,Lu2022}.
Here, we propose an explainable-by-design computational hematology approach underpinned by Vision Language Models (VLM) capable of generating narrative descriptions of cells from PBS and BMA.  By jointly learning from images and text, VLMs associate visual patterns with descriptive language, enabling tasks such as image captioning and visual question answering \cite{ghosh2024exploring}. Recent studies have shown that VLMs can be trained to generate full radiology reports and draft histopathology narratives directly from gigapixel slides \cite{tanno2025collaboration, tran2025generating}. We present HemBLIP, a vision–language modeling approach designed to generate morphology-aware descriptions of white blood cells in the context of leukemia diagnosis from blood film microscopy. We trained our model on a newly constructed morphology-rich dataset of $\sim$14k cell–caption pairs and compared its performance with a medical foundation model (MedGEMMA) under full and Low-Rank Adaptation (LoRA) fine-tuning.

\section{Materials and Methods}
\label{sec:methods}

\subsection{Dataset Construction}
We composed a morphology-aware dataset of peripheral blood smear images combining healthy and leukemic white blood cells (WBCs). For healthy cells, we used the \emph{WBCAtt dataset} \cite{tsutsui2023wbcatt}, containing 10k expert-annotated WBCs (neutrophils, eosinophils, basophils, lymphocytes, monocytes) with 11 categorical attributes describing nuclear, cytoplasmic, and granular features. For leukemia, we used the \emph{LeukemiaAttri dataset} \cite{rehman2024leukemiaattri}, which provides 10k labeled cells across five subtypes (ALL, AML, APML, CLL, CML) and seven expert-defined morphological attributes (e.g., chromatin texture, nucleoli visibility, basophilia). After filtering incomplete or erroneous entries, we selected 7,037 healthy and 7,622 leukemic cells.

%We extract single cells from whole slide images using the provided bounding boxes. 

%following the official train/test partitions from both datasets. - both data sources came with a predefined train/test split (as labels), but agreed, its unnecessary. 
 
%To improve interpretability,

We further paired each cell image with a structured natural-language caption describing its morphology. Feature labels were transformed into templated sentences and augmented with GPT-4 paraphrases while constraining generation to retain factual morphological descriptors and diagnostic labels. The resulting corpus comprises $\sim$14k image–text pairs covering normal and leukemic morphologies. Figure~\ref{fig:qualitative_examples} illustrates two sample images and ground truth generated captions.

\subsection{Model Architectures and Fine-tuning}
%We evaluated two vision–language models (VLMs): a general-purpose model (\emph{BLIP} \cite{li2022blip}) and a biomedical foundation model (\emph{MedGEMMA-4B} \cite{sellergren2025medgemmatechnicalreport}). BLIP integrates a ViT-based vision encoder and transformer text decoder pretrained on web data, while MedGEMMA augments a SigLIP vision tower with a medically tuned decoder trained on diverse clinical images. The architectures can be found in Figure \ref{fig:model_architectures}

%Both models were adapted for blood cell captioning using Low-Rank Adaptation (LoRA) \cite{hu2022lora}. We inserted LoRA adapters into decoder attention layers and fine-tuned them on our image–text pairs, keeping the vision encoders frozen to preserve pretrained representations. For BLIP, we also trained a fully fine-tuned variant to assess domain adaptation vs. parameter-efficient tuning. Each model was trained with AdamW (learning rate $5\times10^{-5}$) and early stopping on validation loss.

\begin{figure*}[t]
    \centering
    \subfloat[\textbf{BLIP architecture overview.} The ViT-based vision encoder and transformer text decoder are aligned through contrastive and captioning objectives.]{
        \includegraphics[width=0.47\textwidth]{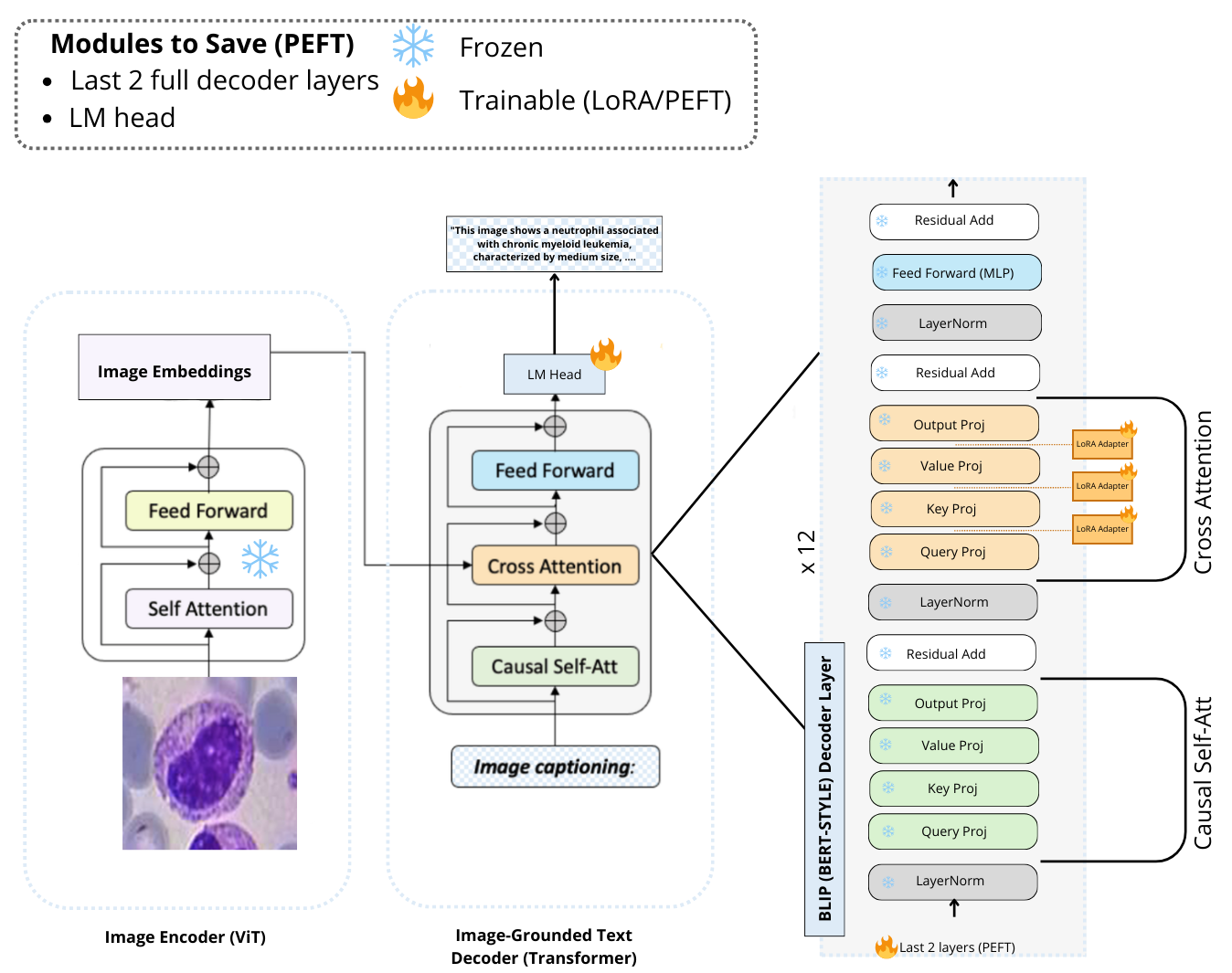}
        \label{fig:blip}
    }
    \hfill
    \subfloat[\textbf{MedGEMMA architecture overview.} A SigLIP vision tower is coupled with a medically tuned decoder trained on multimodal clinical data.]{
        \includegraphics[width=0.47\textwidth]{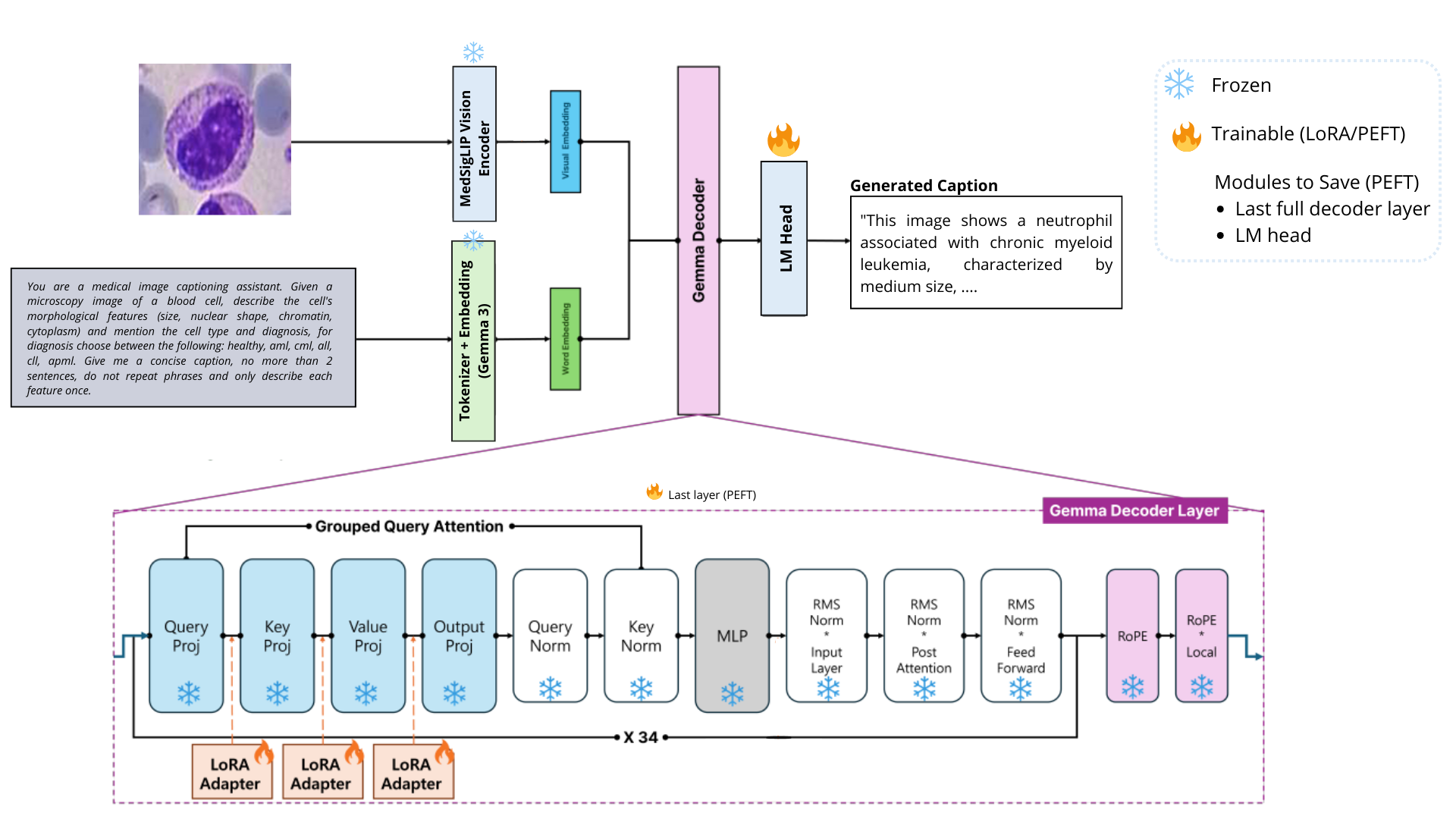}
        \label{fig:medgemma}
    }
    \caption{Comparison of the two vision--language architectures used in this study. Both were fine-tuned on the morphology-aware leukemia dataset using LoRA adapters for parameter-efficient adaptation.}
    \label{fig:model_architectures}
\end{figure*}

We propose HemBLIP, a vision–language model adapted from the general-purpose BLIP architecture \cite{li2022blip} for the task of white blood cell morphology captioning. HemBLIP couples a ViT-based image encoder with a transformer text decoder and is fine-tuned to generate structured, clinically meaningful cell descriptions. To achieve efficient domain adaptation, we apply Low-Rank Adaptation (LoRA) \cite{hu2022lora} to the decoder attention layers while keeping the vision encoder frozen, enabling morphology-aware learning without extensive retraining. We additionally evaluate a fully fine-tuned HemBLIP variant to examine the trade-offs between parameter-efficient and full adaptation strategies.
To contextualize performance, we compared HemBLIP against a biomedical foundation model, MedGEMMA-4B \cite{sellergren2025medgemmatechnicalreport}, which combines a SigLIP vision tower with a medically aligned language decoder trained on diverse clinical image–text data. Model architectures are summarized in Figure \ref{fig:model_architectures}. All models were optimized using AdamW (learning rate $5\times10^{-5}$) with early stopping based on validation loss. The implementation is available at: https://github.com/julievanlogtestijn-ucl/Vision-Language-Models-for-Leukemia-Detection/. 

\subsection{Clinical Relevance Evaluation}
Standard captioning metrics such as BLEU, ROUGE-L, and BERTScore \cite{zhang2020bertscore,lee2020biobert} were computed to assess lexical and semantic similarity with reference captions. To quantify morphological faithfulness, we developed a regex-based attribute extractor that performs controlled string matching over generated captions to identify mentions of predefined cytological attributes (e.g., cell size, chromatin pattern, nucleoli, basophilia). Extracted attributes were compared to ground-truth labels, and confusion matrices were used to evaluate correspondence and categorize biologically plausible errors (e.g., coarse vs. open chromatin, small vs. medium cell size).

To probe whether models encode diagnostically meaningful visual information, we trained lightweight cosine-similarity classifier heads on frozen image embeddings to distinguish between leukemia subtypes and cell types.

\subsection{External Validation}
An additional set of 507 cell images from the Blood Cell Atlas \cite{mindray2022blood}, a public Kaggle collection \cite{singh_bloodcellimages}, and the low-cost microscope subset of the \emph{LeukemiaAttri dataset} \cite{rehman2024leukemiaattri} was used for robustness testing under domain shift. These samples include both healthy and malignant cells imaged under varied staining and sensor conditions, allowing evaluation of real-world generalization.

\section{Results}
\label{sec:results}

\begin{figure}[t]
    \centering
    \includegraphics[width=\columnwidth]{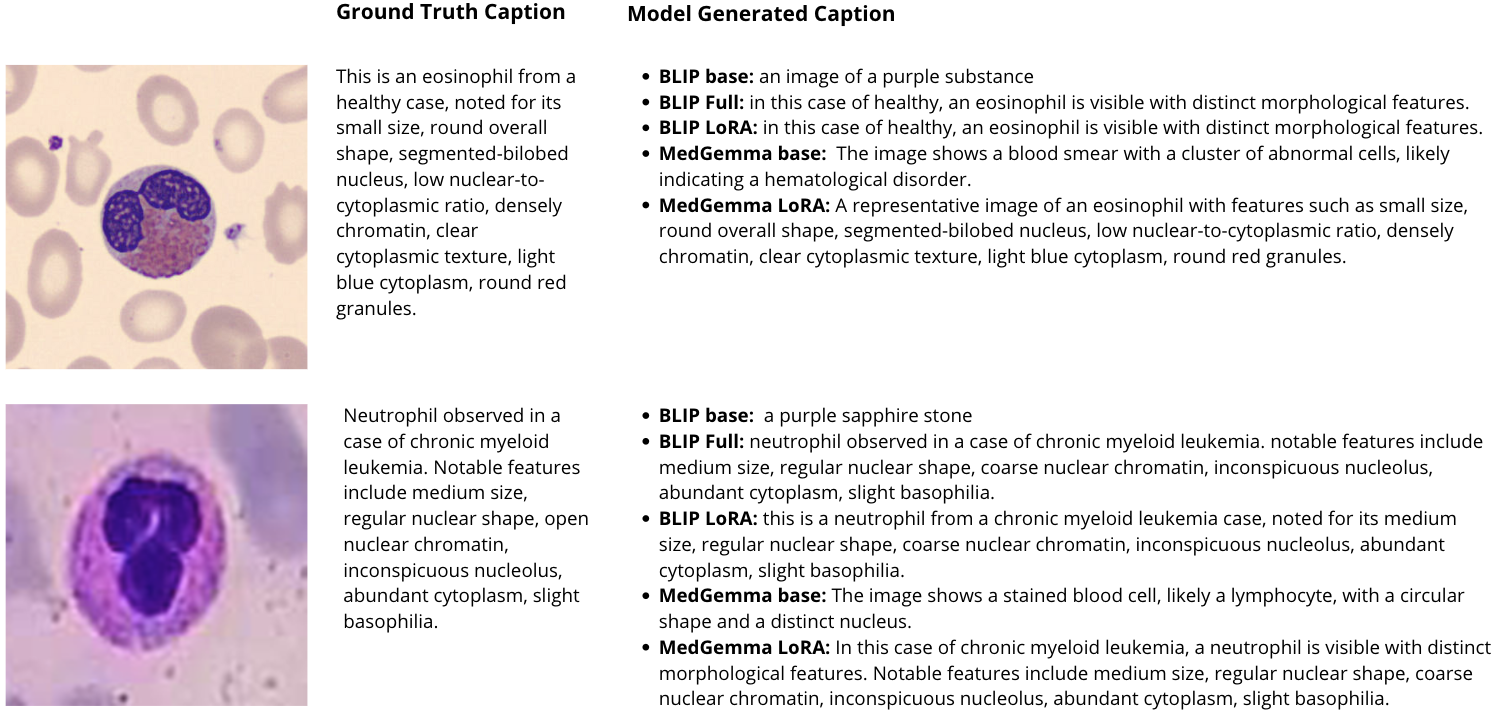}
    \caption{Example blood cell inputs with ground-truth captions and model-generated descriptions. Each model exhibits varying degrees of morphological specificity and diagnostic accuracy.}
    \label{fig:qualitative_examples}
\end{figure}

We evaluated captioning performance, morphological interpretability, and representation quality across internal and external test sets. Example model outputs are shown in Fig.~\ref{fig:qualitative_examples}.

% Results demonstrate that parameter-efficient fine-tuning substantially enhances both caption accuracy and morphological faithfulness while maintaining generalization under domain shift.

\subsection{Caption Generation Performance}
Table~\ref{tab:nlp_metrics_combined} reports quantitative caption generation metrics on both test sets. LoRA-adapted models consistently outperform their fully fine-tuned or base counterparts, confirming the benefits of parameter-efficient adaptation. The domain-specialized \textit{MedGEMMA LoRA} achieves the strongest internal metrics (BLEU~0.31, ROUGE-L~0.52, BERTScore~0.87), while \textit{HemBLIP LoRA}  achieves comparable results with notably lower computational cost. Under external validation, both LoRA variants retain higher semantic alignment (ROUGE-L~0.25, BERTScore~0.78) than their base models, highlighting improved robustness to microscope and staining variations.

\begin{table}[h!]
\centering
\small
\caption{Caption generation metrics on internal and external test sets.}
\label{tab:nlp_metrics_combined}
\resizebox{\columnwidth}{!}{%
\begin{tabular}{lcccccc}
\toprule
 & \multicolumn{3}{c}{\textbf{Internal}} & \multicolumn{3}{c}{\textbf{External}} \\
\cmidrule(lr){2-4}\cmidrule(lr){5-7}
\textbf{Model} & \textbf{BLEU} & \textbf{ROUGE-L} & \textbf{BERTScore F1} & \textbf{BLEU} & \textbf{ROUGE-L} & \textbf{BERTScore F1} \\
\midrule
HemBLIP Full & 0.24 & 0.42 & 0.83 & 0.02 & 0.22 & 0.77 \\
HemBLIP LoRA & 0.27 & 0.49 & 0.86 & \textbf{0.02} & \textbf{0.25} & \textbf{0.78} \\
MedGEMMA Base & 0.02 & 0.13 & 0.74 & 0.02 & 0.18 & 0.74 \\
MedGEMMA LoRA & \textbf{0.31} & \textbf{0.52} & \textbf{0.87} & 0.02 & 0.23 & 0.77 \\
\bottomrule
\end{tabular}%
}
\end{table}

\subsection{Morphological Attribute Accuracy}
To assess the relevance of descriptions, we extracted key morphological attributes (e.g., cell size, chromatin texture, nuclear shape) from generated captions using a regex-based evaluator. As shown in Table~\ref{tab:feature_accuracy}, HemBLIP models consistently produced more accurate and clinically relevant morphological attributes than MedGEMMA variants, both internally and externally. While LoRA fine-tuning yielded only minor gains over full HemBLIP, it provided comparable performance at substantially lower training cost. All models exhibited expected degradation under external evaluation, reflecting domain shift effects. 

\begin{table*}[h!]
\centering
\small
\caption{Feature-level accuracy (\%) for morphological attributes extracted from generated captions on internal and external test sets. Only frequently mentioned features are shown.}
\label{tab:feature_accuracy}
\resizebox{\textwidth}{!}{%
\begin{tabular}{lcccccccc}
\toprule
\textbf{Feature} & \multicolumn{2}{c}{\textbf{HemBLIP Full}} & \multicolumn{2}{c}{\textbf{HemBLIP LoRA}} & \multicolumn{2}{c}{\textbf{MedGEMMA Base}} & \multicolumn{2}{c}{\textbf{MedGEMMA LoRA}} \\
\cmidrule(lr){2-3}\cmidrule(lr){4-5}\cmidrule(lr){6-7}\cmidrule(lr){8-9}
 & \textbf{Int.} & \textbf{Ext.} & \textbf{Int.} & \textbf{Ext.} & \textbf{Int.} & \textbf{Ext.} & \textbf{Int.} & \textbf{Ext.} \\
\midrule
Cell type & 73.26 & 38.60 & 72.40 & 29.58 & 14.70 & 7.93 & 57.73 & 22.45 \\
Nuclear chromatin texture & 52.72 & 32.41 & 55.47 & 31.12 & 51.93 & 42.59 & 59.30 & 31.36 \\
Cytoplasm amount & 70.35 & 50.23 & 69.87 & 43.91 & 26.67 & 0.00 & 57.37 & 35.99 \\
Diagnosis & 79.14 & 43.45 & 82.27 & 53.85 & 16.96 & 21.74 & 54.09 & 41.03 \\
Nuclear shape & 77.38 & 7.91 & 72.90 & 14.25 & 56.62 & 65.68 & 64.22 & 12.06 \\
Overall shape & 77.44 & 11.46 & 77.68 & 15.78 & 61.79 & 14.99 & 75.80 & 17.79 \\
Cell size & 91.80 & 47.78 & 85.37 & 41.07 & 32.33 & 19.90 & 81.15 & 31.23 \\
\bottomrule
\end{tabular}%
}
\end{table*}

\subsection{HemBLIP as a feature extractor}
Finally, we evaluated whether visual encoders captured discriminative morphology by training lightweight classifiers on frozen embeddings from each backbone. Since LoRA fine-tuning did not modify the vision encoders, this comparison focuses on the \emph{HemBLIP Full} model (fully fine-tuned, including the vision tower), the original \emph{BLIP Base}, the \emph{MedGEMMA} backbone, and a random projection baseline. 

As shown in Table~\ref{tab:classification_combined}, HemBLIP Full achieved the strongest performance for both leukemia subtype (Acc.~0.85, F1~0.83; 5 classes) and cell-type classification (Acc.~0.60, F1~0.64; 13 classes), outperforming MedGEMMA and baseline embeddings. External test performance dropped substantially (Acc.~0.52 / 0.33), highlighting ongoing challenges in domain generalization. These findings suggest that full HemBLIP fine-tuning yields visual representations that retain clinically meaningful diagnostic cues beyond caption generation.

\begin{table}[h!]
\centering
\small
\caption{Frozen-backbone classifier performance on internal and external test sets. Accuracy and weighted F1 scores are reported for leukemia subtype and cell-type classification.}
\label{tab:classification_combined}
\resizebox{\columnwidth}{!}{%
\begin{tabular}{llcccc}
\toprule
\multirow{2}{*}{\textbf{Task}} & \multirow{2}{*}{\textbf{Backbone}} 
& \multicolumn{2}{c}{\textbf{Internal}} 
& \multicolumn{2}{c}{\textbf{External}} \\
\cmidrule(lr){3-4}\cmidrule(lr){5-6}
 &  & \textbf{Acc.} & \textbf{F1} & \textbf{Acc.} & \textbf{F1} \\
\midrule
\multirow{4}{*}{Leukemia subtype} 
 & HemBLIP Full     & \textbf{0.848} & \textbf{0.834} & \textbf{0.517} & 0.474 \\
 & BLIP Base                   & 0.807 & 0.804 & 0.501 & \textbf{0.516} \\
 & MedGEMMA            & 0.708 & 0.713 & 0.456 & 0.474 \\
 %& Random projection baseline  & 0.668 & 0.678 & 0.326 & 0.223 \\
\midrule
\multirow{4}{*}{Cell type} 
 & HemBLIP Full   & \textbf{0.595} & \textbf{0.644} & \textbf{0.330} & \textbf{0.363} \\
 & BLIP Base                   & 0.520 & 0.570 & 0.151 & 0.183 \\
 & MedGEMMA            & 0.344 & 0.353 & 0.117 & 0.132 \\
 %& Random projection baseline  & 0.348 & 0.363 & 0.169 & 0.165 \\
\bottomrule
\end{tabular}%
}
\end{table}

\section{Conclusion}
\label{sec:conclusion}

This work presented an explainable AI framework for leukemia diagnosis from blood films using vision–language models. By constructing a morphology-rich dataset that combines expert annotations with GPT-generated captions, we enable models to learn descriptive reasoning aligned with hematological practice. Using parameter-efficient LoRA fine-tuning, both a general-domain model (BLIP) and a biomedical model (MedGEMMA) were adapted to generate morphology-aware captions and diagnostic labels.

Our results show that LoRA adaptation not only reduces computational cost but also improves the clinical relevance of generated explanations. Our HemBLIP model produced the most accurate and interpretable descriptions, demonstrating that general-purpose VLMs can be successfully repurposed for specialized medical images. The proposed evaluation pipeline—linking textual outputs to morphological attributes and testing discriminative power—provides a practical approach to assess explainability in medical AI.

Overall, this study highlights how morphology-aware VLMs can bridge the gap between automated image analysis and human interpretability, offering a scalable path toward transparent, clinically useful AI-assisted leukemia diagnostics.

\section{Acknowledgments}
No funding was received for conducting this study. The authors have no relevant financial or non-financial interests to disclose.

\section{Compliance with Ethical Standards}
This research did not involve any studies with human participants or animals performed by any of the authors. The study used only publicly available datasets; therefore, ethical approval was not required.

%\vfill
%pagebreak

\bibliographystyle{IEEEbib}
\bibliography{refs}

\end{document}